\documentclass[sigconf]{acmart}

\usepackage{array}
\usepackage[table]{xcolor}       % 表格着色
\usepackage{booktabs}
\usepackage{tabularx}
\usepackage{multirow}
\usepackage{adjustbox}
\usepackage[most]{tcolorbox}

\usepackage{enumitem}            % 如需严格控制列表格式可保留，否则建议注释
\usepackage{fontawesome}
\usepackage{microtype}

\usepackage{listings}
\usepackage{stfloats}
\usepackage{caption}
\AtBeginDocument{%
  }

\copyrightyear{2026}
\acmYear{2026}
\setcopyright{cc}
\setcctype{by}
\acmConference[KDD '26]{Proceedings of the 32nd ACM SIGKDD Conference on Knowledge Discovery and Data Mining V.2}{August 09--13, 2026}{Jeju Island, Republic of Korea}
\acmBooktitle{Proceedings of the 32nd ACM SIGKDD Conference on Knowledge Discovery and Data Mining V.2 (KDD '26), August 09--13, 2026, Jeju Island, Republic of Korea}
\acmDOI{10.1145/3770855.3817552}
\acmISBN{979-8-4007-2259-2/2026/08}
\begin{document}

%%
%% The "title" command has an optional parameter,
%% allowing the author to define a "short title" to be used in page headers.
%\title{\texttt{OmniPhys}: Knowledge-Graph-Driven Benchmarking and Collective Optimization for Physical Commonsense in Text-to-Image Generation}
\title[OmniPhys: Knowledge-Graph-Driven Benchmarking and Collective Optimization for Physical Commonsense \\ in Text-to-Image Generation]{\texttt{OmniPhys}: Knowledge-Graph-Driven Benchmarking and Collective Optimization for Physical Commonsense in Text-to-Image Generation}

%%
%% The "author" command and its associated commands are used to define
%% the authors and their affiliations.
%% Of note is the shared affiliation of the first two authors, and the
%% "authornote" and "authornotemark" commands
%% used to denote shared contribution to the research.
\author{Yajing Xu}
% \authornote{Both authors contributed equally to this research.}
%\email{yajingxu@zju.edu.cn}
%\orcid{1234-5678-9012}
% \author{G.K.M. Tobin}
% \authornotemark[1]
% \email{webmaster@marysville-ohio.com}
\affiliation{%
  \institution{Zhejiang University}
  \city{Hangzhou, Zhejiang}
  %\state{Ohio}
  \country{China}
}

\email{yajingxu@zju.edu.cn}

\author{Yarong Lan}
\affiliation{%
  \institution{Zhejiang University}
  \city{Hangzhou, Zhejiang}
  \country{China}}
\email{yrlan16@zju.edu.cn}

\author{Jiaoyan Chen}
\affiliation{%
  \institution{The University of Manchester}
  \city{Manchester}
  \country{United Kingdom}
}
\email{jiaoyan.chen@manchester.ac.uk}

\author{Yichi Zhang}
\affiliation{%
  \institution{Zhejiang University}
  \city{Hangzhou,   Zhejiang}
  \country{China}
}
\email{zhangyichi.each@zju.edu.cn}

\author{Jeff Z. Pan}
\affiliation{%
 \institution{The University of Edinburgh}
 \city{Edinburgh}
 \country{United Kingdom}}
\email{j.z.pan@ed.ac.uk}

\author{Mingchen Tu}
\affiliation{%
  \institution{Zhejiang University}
  \city{Ningbo, Zhejiang}
  \country{China}}
\email{mingchentz@zju.edu.cn}

\author{Zhizhen Liu}
\affiliation{%
  \institution{Ant Group}
  \city{Hangzhou, Zhejiang}
  \country{China}}
\email{zhizhen.lzz@antgroup.com}

\author{Wen Zhang}
\authornote{Corresponding Author}
\affiliation{%
  \institution{Zhejiang University}
  \city{Hangzhou, Zhejiang}
  \country{China}}
\email{zhang.wen@zju.edu.cn}

\author{Huajun Chen}
\authornotemark[1]
\affiliation{%
  \institution{Zhejiang University}
  \city{Hangzhou, Zhejiang}
  \country{China}}
\email{huajunsir@zju.edu.cn}

\renewcommand{\shortauthors}{Yajing Xu et al.}
%%
%% By default, the full list of authors will be used in the page
%% headers. Often, this list is too long, and will overlap
%% other information printed in the page headers. This command allows
%% the author to define a more concise list
%% of authors' names for this purpose.
%%
%% The abstract is a short summary of the work to be presented in the
%% article.
\begin{abstract}
While text-to-image models exhibit remarkable visual fidelity, they frequently violate fundamental physical commonsense. Existing benchmarks often rely on coarse-grained descriptions, failing to diagnose the mastery of specific physical principles. Moreover, the high stochasticity of generative processes causes current prompt optimization methods to suffer from "gradient hallucinations," where optimizers are misled by transient visual artifacts rather than systemic flaws. To address these challenges, we introduce \textbf{OmniPhys}, a rigorous benchmark of 1,551 samples grounded in a Physical Knowledge Graph. By aligning PhET simulations with standard curricula, OmniPhys operationalizes a knowledge-to-scenario pipeline that performs diagnostic stress tests via a dual-path verification protocol. We further propose \textbf{OmniPrompt}, an iterative framework that treats physical alignment as a discrete optimization problem. For each query, OmniPrompt aggregates $K$ stochastic images into a per-query feedback buffer. Across training, it further merges feedback from batches of $B$ queries before each meta-policy update, filtering seed and query-local noise. Evaluations across 12 representative T2I models reveal universal physical bottlenecks. Results demonstrate that OmniPrompt significantly enhances physical consistency across diverse backbones, proving the transferability and efficacy of our evolved meta-policies. The code and data are available at \url{https://github.com/zjukg/OmniPhys}
\end{abstract}

%%
%% The code below is generated by the tool at http://dl.acm.org/ccs.cfm.
%% Please copy and paste the code instead of the example below.
%%


%%
%% Keywords. The author(s) should pick words that accurately describe
%% the work being presented. Separate the keywords with commas.
%% A "teaser" image appears between the author and affiliation
%% information and the body of the document, and typically spans the
%% page.

% \received{20 February 2007}
% \received[revised]{12 March 2009}
% \received[accepted]{5 June 2009}

%%
%% This command processes the author and affiliation and title
%% information and builds the first part of the formatted document.
\maketitle
\begin{figure}[t]  % [t]=top, [b]=bottom, [h]=here, [!htbp] 更灵活
    \centering
    \includegraphics[width=0.85\linewidth]{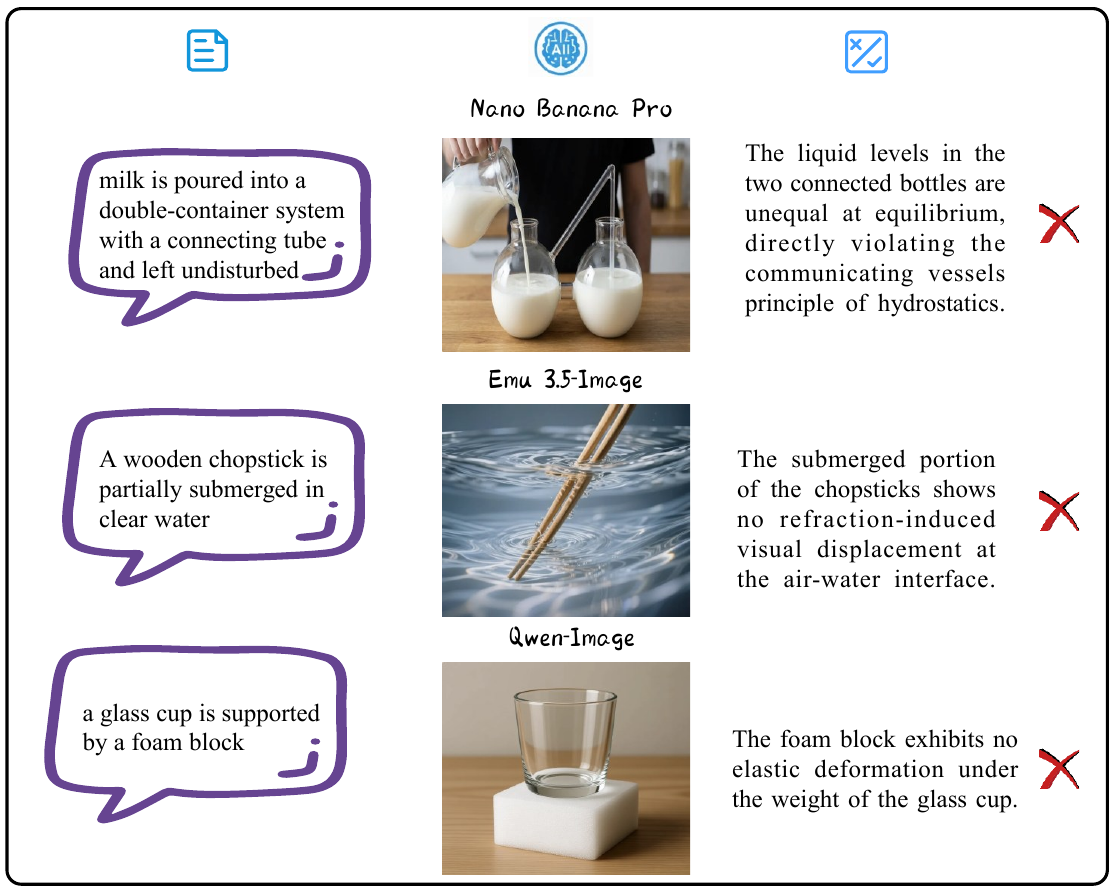}
    \caption{Examples of physical commonsense violations across frontier T2I models, illustrating failures in 
    hydrostatics (top, communicating vessels principle), optics  (middle, light refraction), and mechanics  (bottom, elastic deformation under stress).}
    \label{fig:example}
\end{figure}
\section{Introduction}
The rapid advancement of Text-to-Image (T2I) models has revolutionized digital content creation, enabling the generation of high-fidelity images from complex natural language descriptions \cite{DBLP:conf/nips/SahariaCSLWDGLA22,DBLP:conf/cvpr/RombachBLEO22, DBLP:conf/icml/EsserKBEMSLLSBP24, DBLP:journals/corr/abs-2505-22705, DBLP:journals/corr/abs-2503-21758}. However, a critical ``logical ceiling'' has emerged as these models transition from artistic tools to simulators of reality: the persistent violation of physical commonsense. As illustrated in Figure ~\ref{fig:example}, even frontier models fail to respect fundamental laws of nature. Examples include unequal liquid levels in communicating vessels, missing refraction-induced displacement at air-water interfaces, and the absence of elastic deformation under mechanical stress. These ``physical hallucinations''~\cite{PRKS+2023} reveal that current models primarily rely on superficial pattern recognition rather than an internalized understanding of world physics.

Evaluating and rectifying these violations presents two formidable challenges. First, existing benchmarks lack the granularity and scale required for rigorous diagnostics. Early efforts like Commonsense-T2I ~\cite{DBLP:journals/corr/abs-2406-07546} and PhyBench ~\cite{DBLP:journals/corr/abs-2406-11802} rely on phenomenological scenarios and provide only coarse-grained analysis at the domain level. They may identify general failures in mechanics, but fail to pinpoint specific knowledge gaps. Consequently, they cannot diagnose whether a model lacks mastery of foundational laws, such as Archimedes' principle or the leverage principle. Second, aligning T2I models with physical constraints presents a unique optimization dilemma. Conventional Supervised Fine-Tuning (SFT) \cite{DBLP:conf/nips/HaoC0W23,DBLP:conf/aaai/HeiGWWWZ24} and Reinforcement Learning \cite{DBLP:journals/corr/abs-2305-16381,DBLP:conf/nips/XuLWTLDTD23,DBLP:conf/aaai/AnZLZFHC0P25} methods are often prohibitively resource-intensive, necessitating both substantial computational overhead and extensive curated datasets of high-fidelity image-text pairs.
While discrete prompt optimization provides a lightweight alternative using Large Language Models (LLMs), current methods \cite{DBLP:conf/emnlp/PryzantI0L0023,DBLP:conf/iclr/Yang0LLLZC24,DBLP:conf/acl/YanWZZWXLKK25} primarily focus on direct text-to-text interaction. In T2I scenarios, the LLM acts as an intermediary agent. The optimization goal shifts to improving the output of a secondary system (the T2I model) rather than the LLM itself. This setup is plagued by the high stochasticity of the T2I generation manifold. Consequently, the optimizer often suffers from ``gradient hallucinations'': misleading textual gradients caused by (i) transient visual artifacts from a single stochastic sample and/or (ii) query-local failures that do not generalize across the training pool.

To overcome these hurdles, we first introduce \textbf{OmniPhys} (short for \textbf{Omni}-category \textbf{Phys}ics), a rigorous benchmark comprising 1,551 curated samples anchored to 14 Physical Knowledge Points (PKPs) organized under three domains (Mechanics, Optics, and Object Properties). Its construction operationalizes a \textbf{knowledge-to-scenario} pipeline grounded in a hierarchical Physical Knowledge Graph (PKG)~\cite{PVGW2017}. To ensure physical rigor, we align PhET \footnote{https://phet.colorado.edu/} Interactive Simulations with standard physics curricula to extract foundational principles across mechanics, optics, and object properties. Our evaluation employs implicit queries that describe a physical scene without explicitly stating the outcome, such as ``an iron block is placed in a container of water''. By anchoring each query to a target PKP and its PKG-annotated atomic physical statements, OmniPhys provides a diagnostic ``stress test'' of a model's authentic reasoning. This design forces the model to derive the correct physical manifestation from its internalized knowledge rather than simple keyword matching. We also implement a Strict  \textbf{Dual-Path Physical Verification Protocol}, which combines discriminative VQA probes with descriptive consistency auditing. By requiring a model to both correctly identify the physical outcome and maintain structural consistency in its visual execution, this protocol effectively filters out heuristic-based ``lucky guesses" and provides a high-fidelity diagnostic of the model’s physical alignment.

Building on this diagnostic foundation, we further propose \textbf{OmniPrompt}, an iterative framework that transforms physical alignment into a stable, closed-loop optimization process. To mitigate stochastic generation noise, \textbf{OmniPrompt} uses a two-level aggregation scheme.
At the query level, it consolidates audits over an ensemble of $K$ images into a feedback buffer $e_i$.
At the optimization level, it merges buffers from a batch of $B$ queries before deriving linguistic gradients and updating the meta-policy, so that updates reflect shared physical failures rather than a single query or seed. Consequently, the system evolves high-level instructions that are generalized to diverse physical domains.

In summary, our primary contributions are as follows:
\begin{itemize}[leftmargin=*,nosep]
    \item \textbf{Systematic Diagnostics:} We release \textbf{OmniPhys}, the first T2I benchmark grounded in a Physical Knowledge Graph. Its taxonomy aligns PhET simulations with standard physics curricula, enabling fine-grained evaluation across 14 PKPs under three domains.
    \item \textbf{Comprehensive Model Analysis:} We conduct an extensive study across 12 representative T2I models, including both frontier closed-source engines and diverse open-source architectures. This provides a granular landscape of current physical reasoning capabilities and identifies universal bottlenecks such as optical reflection and structural mechanics.
    \item \textbf{Robust and Transferable Optimization:} We propose \textbf{OmniPrompt}, which 
    addresses ``gradient hallucinations'' by batch-merging per-query feedback over $B$ training queries before each meta-policy update via textual gradient descent. We demonstrate its efficacy across diverse backbones, proving that our evolved meta-policies provide significant and transferable performance gains. 
\end{itemize}
\section{Related Work}
\subsection{Advancements in Text-to-Image Models}
The landscape of T2I synthesis has been recently redefined by a surge in high-fidelity generative capabilities, spanning from stylized art to complex photorealism. Driven by breakthroughs in diffusion processes \cite{DBLP:conf/icml/EsserKBEMSLLSBP24,DBLP:journals/corr/abs-2511-22699,DBLP:journals/corr/abs-2505-22705}, autoregressive modeling \cite{DBLP:conf/cvpr/HanL0YZYPL25,DBLP:journals/corr/abs-2503-21758} and multimodal unified system \cite{DBLP:journals/corr/abs-2501-17811,DBLP:journals/corr/abs-2510-26583}, modern T2I systems can produce stunning visual content; however, their outputs often betray a lack of physical groundedness. Instead of manifesting a principled understanding of reality, these models frequently default to probabilistic associations, leading to conspicuous failures in maintaining structural integrity or fluidic realism. To bridge this gap, we introduce a novel optimization framework that moves beyond mere visual alignment, iteratively refining generative policies to ensure adherence to essential physical laws.
\subsection{Evaluation of Text-to-Image Models}
Evaluating T2I models remains a multi-faceted challenge. Traditional metrics like FID \cite{DBLP:conf/nips/HeuselRUNH17} focus on visual quality, while CLIP-based metrics \cite{DBLP:conf/emnlp/HesselHFBC21} measure semantic alignment. To capture higher-order capabilities, benchmarks such as ABC-6K \cite{DBLP:conf/iclr/FengHFJANBWW23} and T2I-CompBench \cite{DBLP:conf/nips/HuangSXLL23} have shifted focus toward compositional reasoning, including spatial relations and attribute binding. More recently, WorldGenBench \cite{DBLP:journals/corr/abs-2505-01490} and WISE \cite{DBLP:journals/corr/abs-2503-07265} have explored generation grounded in world knowledge, while R2I-Bench \cite{DBLP:conf/emnlp/ChenLXSYRZH25} targets textual commonsense. Despite these efforts, physical commonsense remains under-explored. Initial attempts like Commonsense-T2I ~\cite{DBLP:journals/corr/abs-2406-07546} and PhyBench ~\cite{DBLP:journals/corr/abs-2406-11802} categorize specific physical errors but lack a systematic taxonomy. \textbf{Unlike prior work, OmniPhys provides a knowledge-graph-driven taxonomy and a strict dual-path verification protocol for fine-grained physical diagnosis; we further propose OmniPrompt, an iterative meta-policy optimization framework built on this diagnostic signal.}

\begin{figure*}[t]  % [t]=top, [b]=bottom, [h]=here, [!htbp] 更灵活
    \centering
    \includegraphics[width=\linewidth, height=0.35\textheight, keepaspectratio]{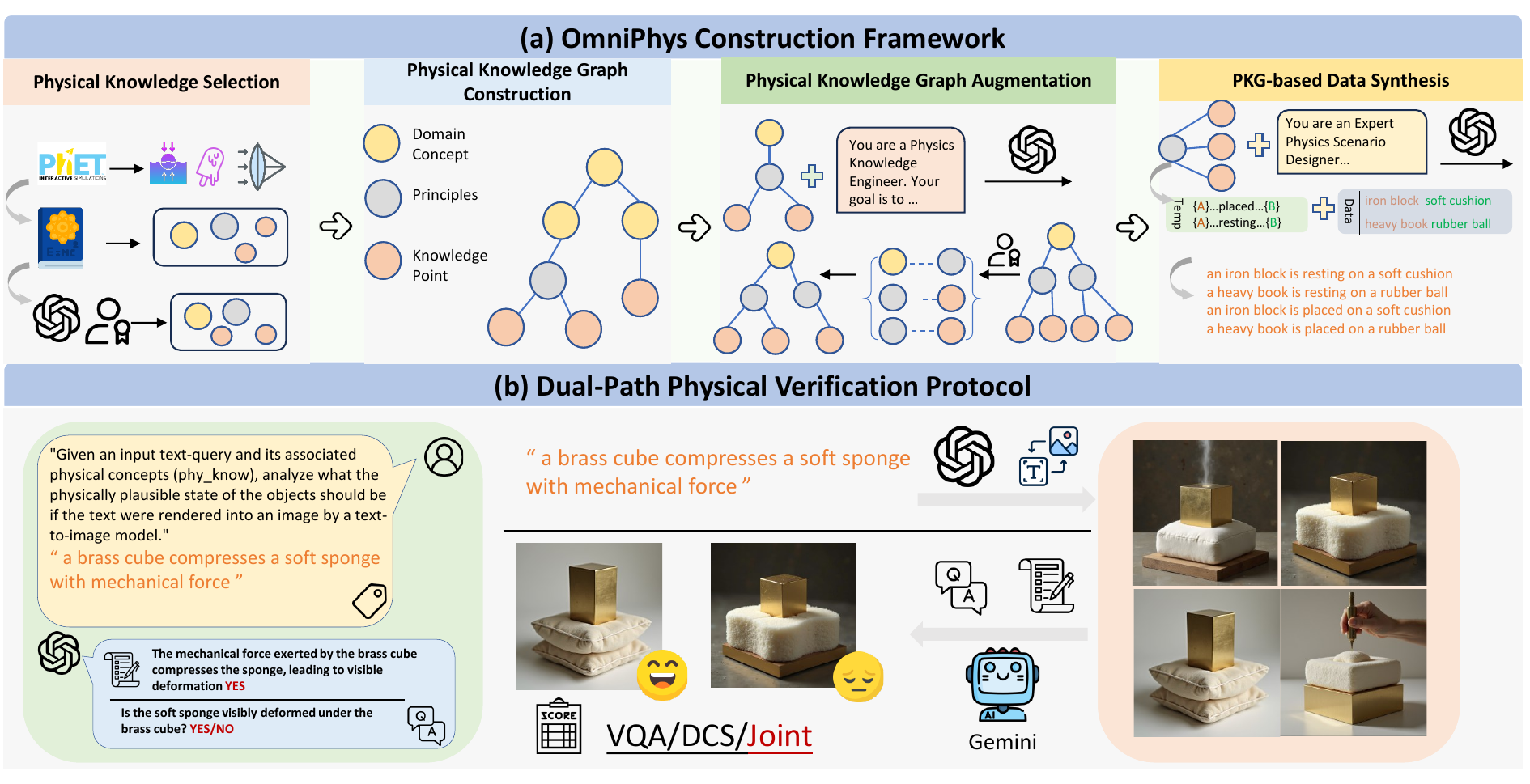}
    \caption{OmniPhys Infrastructure. (a) Construction: A hierarchical pipeline mapping PhET simulations to a PKG for synthesizing high-fidelity prompts. (b) Verification: a dual-path engine that cross-checks discriminative VQA probes against descriptive consistency statements (DCS).}
    \label{fig:bench_framework}
\end{figure*}
\section{The \texttt{OmniPhys} Benchmark}
\label{sec:benchmark}
\textbf{OmniPhys} is a knowledge-graph-driven benchmark designed to address the systematic rigor limitations of existing physical commonsense benchmarks ~\cite{DBLP:journals/corr/abs-2406-07546,DBLP:journals/corr/abs-2406-11802}. As illustrated in Figure \ref{fig:bench_framework}, the benchmark consists of two synergetic modules: (1) a \textbf{Construction Framework} that synthesizes high-fidelity, implicit physical queries grounded in a Physical Knowledge Graph; and (2) a \textbf{Dual-Path Physical Verification Protocol} that serves as an automated diagnostic engine. Unlike traditional holistic scoring, this protocol decomposes physical consistency into PKG-grounded coupled probes including discriminative visual questions (VQA) and descriptive consistency statements (DCS), thereby ensuring that evaluation is both granular and objective.

\subsection{OmniPhys Construction Framework}
\label{sec:construction}

As illustrated in Figure \ref{fig:bench_framework} (a), our pipeline ensures high-fidelity physical alignment and visual verifiability through the following four distinct stages.

\noindent\textbf{(1) Physical Knowledge Selection.}
To ensure physical rigor, we ground our taxonomy in the PhET Interactive Simulations. We align PhET domains with standard physics curricula to extract principles across mechanics, optics and object properties. Subsequently, we use LLM to filter out abstract concepts (e.g., entropy) that lack unambiguous visual signatures, retaining only those with observable binary states (e.g., sinking vs. floating).

\noindent\textbf{(2) Physical Knowledge Graph Construction.}
We formalize the curated principles into a PKG, defined as $\mathcal{G} = (\mathcal{V}, \mathcal{E})$, which serves as the deterministic backbone for data synthesis and evaluation grounding:
\begin{itemize}[leftmargin=*]
    \item \textbf{Node types $\mathcal{V}$:}
    \textit{Domain} (e.g., Object Properties), \textit{Domain Concept} (e.g., density), \textit{Principle} (e.g., Archimedes' principle), and leaf \textit{Physical Knowledge Points (PKPs)}.
    Each PKP anchors one evaluable topic and is annotated with one or more \textbf{atomic physical statements}---short, visually verifiable propositions of expected outcomes under the corresponding principle (e.g., float vs.\ sink vs.\ suspend under density comparison).
    These annotations supply PKG-grounded knowledge for synthesis and probe generation; they are not used verbatim as evaluation prompts.
    \item \textbf{Semantic Relations $\mathcal{E}$:} Directed edges typed as \texttt{is-a} (Domain $\rightarrow$ Concept) and \texttt{derives} (Concept/Principle $\rightarrow$ Principle/PKP, or Concept $\rightarrow$ PKP when no Principle is defined).
    Atomic statements are stored as node-level annotations on PKPs and are therefore not represented as separate graph nodes or edges.
\end{itemize}
Full PKG statistics, including the number of PKPs and atomic statements, are reported in Section~\ref{sec:pkg-stats}.

\noindent\textbf{(3) Physical Knowledge Graph Augmentation.}
To improve the completeness of the PKG in covering visualizable physics, we implement an LLM-based augmentation mechanism. Specifically, using each Domain Concept node as a root, we extract its directly associated Principles and PKPs to form contextual prompts for an LLM. The LLM is tasked with generating candidate PKPs within that domain that are not yet included but remain visually verifiable. Generated candidates are filtered for physical accuracy, logical consistency, and image representability before integration into the graph.

\noindent\textbf{(4) PKG-based Data Synthesis.}
This stage synthesizes evaluation samples anchored to specific PKPs in the PKG, using the PKP's atomic physical statements as knowledge constraints while deliberately hiding the expected outcome:
\begin{itemize}[leftmargin=*,nosep]
    \item \textbf{Scenario Templating:} For each target PKP, LLMs generate diverse scene templates that establish a premise without revealing the result, e.g., \textit{``a \{object\} is placed into a container of \{liquid\}''}.
    \item \textbf{Knowledge-Driven Slot Filling:} Based on the material and state constraints encoded in the PKP's atomic statements, compatible entities are selected to create a clear physical contrast (e.g., \{object\}: iron block; \{liquid\}: water).
    \item \textbf{Implicit Synthesis:} These components are combined into final implicit prompts that omit the expected physical outcome, such as \textit{``an iron block is placed into a container of water''}. By excluding outcome-revealing keywords (e.g., ``sinking''), the prompt forces physically accurate generation to rely on internalized knowledge rather than textual cues.
\end{itemize}

\subsubsection{PKG Scale and Coverage}
\label{sec:pkg-stats}

The finalized PKG is $\mathcal{G}=(\mathcal{V},\mathcal{E})$ with $|\mathcal{V}|=33$ nodes and $|\mathcal{E}|=32$ directed edges, organized into four layers---\textit{Domain} (3), \textit{Domain Concept} (7), \textit{Principle} (9), and \textit{PKP} (14)---with 17 atomic physical statements in total (Table~\ref{tab:pkg-coverage}).
Mechanics spans buoyancy (solid--fluid, gas, immiscible liquids), leverage, and pressure (including communicating vessels); Object Properties covers mass--volume--density relations and melting; Optics covers propagation, reflection, refraction, diffraction, and interference.
Every PKP satisfies visual verifiability: each atomic statement admits at least one binary observable outcome in image space (e.g., float vs.\ sink).

\begin{table}[t]
\centering
\small
\caption{PKG coverage by domain. Edge counts include all \texttt{is-a} and \texttt{derives} relations within each domain.}
\label{tab:pkg-coverage}
\begin{tabular}{lccccc}
\toprule
\textbf{Domain} & \textbf{Concept} & \textbf{Principle} & \textbf{PKP} & \textbf{Edge} & \textbf{Stmt.} \\
\midrule
Mechanics          & 3 & 2 & 6 & 11 & 9  \\
Object Properties  & 3 & 2 & 3 & 10 & 3  \\
Optics             & 1 & 5 & 5 & 11 & 5  \\
\midrule
\textbf{Total}     & \textbf{7} & \textbf{9} & \textbf{14} & \textbf{32} & \textbf{17} \\
\bottomrule
\end{tabular}
\end{table}

\begin{figure}[t]  % [t]=top, [b]=bottom, [h]=here, [!htbp] 更灵活
    \centering
    \includegraphics[width=0.95\linewidth]{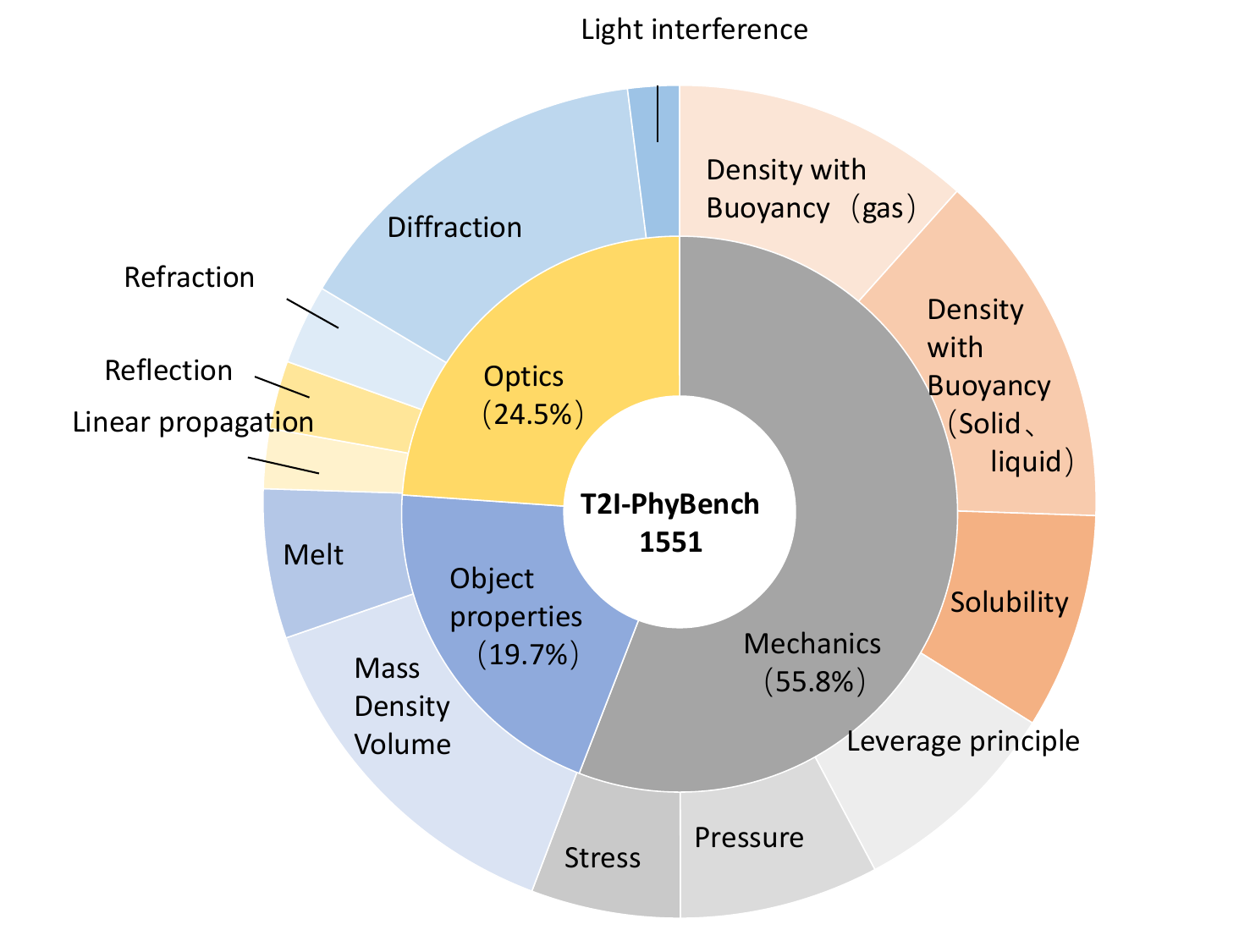}
    \caption{\textbf{OmniPhys Taxonomy.} Hierarchical distribution of 14 PKPs across three domains: Mechanics, Optics, and Object Properties.}
    \vspace{-5mm}
    \label{fig:physbench_stats}
\end{figure}

\paragraph{PKG-to-benchmark linkage.}
OmniPhys samples are synthesized from PKG PKPs. Each sample is anchored to one PKP and includes an implicit prompt $x$, the PKP's atomic physical statements, and LLM-generated coupled probes $(\text{VQA}_i, \text{DCS}_i)_{i=1}^{n}$, where $n$ is sample-specific.
The benchmark contains $N=1{,}551$ implicit prompts in total.
All 14 PKPs are represented in the released benchmark (100\% PKP coverage).

\subsection{Dual-Path Physical Verification Protocol}
\label{sec:dual_path_protocol}
Evaluating physical commonsense requires distinguishing between authentic reasoning and accidental artifacts. We propose \textbf{Dual-Path Convergence}: an image is deemed physically sound only if it passes (1) a \textbf{Discriminative Path} (binary VQA verification) and (2) a \textbf{Descriptive Path} (alignment with LLM-generated DCS probes).
\paragraph{(1) PKG-Guided Coupled Probe Generation.}
For each implicit prompt $x$, we retrieve the atomic physical statements of its anchor PKP and prompt an LLM with $(x, \{\text{atomic statements}\})$ to generate $n$ coupled units
$u_i=(\text{VQA}_i, \text{DCS}_i)_{i=1}^{n}$.
Here, $n$ is sample-specific: it denotes the physical state points that the LLM selects to examine for $x$ under PKG constraints, rather than the number of PKG atomic statements.
Each unit pairs one discriminative visual question with a matching descriptive consistency statement.
Probe generation follows three pillars:
\begin{itemize}[leftmargin=*, nosep]
    \item \textbf{Visual-Physical Meta-Template}: Restricts output to observable, static outcomes while pruning transient processes or invisible quantities unjudgeable from 2D projections.
    \item \textbf{Categorized Few-Shot Logic}: Aligns probe generation with the reasoning patterns of the 14 PKG PKPs organized under three domains via expert-curated demonstrations.
    \item \textbf{Coupled Structure}: For each LLM-selected physical state point, the generator co-produces a VQA probe and a matching DCS statement.
\end{itemize}
Following this PKG-guided pipeline (see Appendix ~\ref{app:probe-generation}), the generator outputs structured JSON arrays of coupled $(\text{VQA}_i, \text{DCS}_i)_{i=1}^{n}$ pairs with gold answers.

\paragraph{(2) Automated Scoring and Evaluation.} We employ a Vision-Language Model (VLM) as an automated evaluator. For each prompt $x_j$, we generate 4 independent images $\{G_{j,k}\}_{k=1}^4$. Let $n_j$ denote the number of coupled probes generated for prompt $x_j$. For each image $G_{j,k}$, we adopt a stringent ``all-or-nothing'' policy:
\begin{itemize}[leftmargin=*, nosep]
    \item \textbf{VQA Score}: $S_{VQA}(G_{j,k}) = 1$ if the VLM's responses to all $n_j$ generated discriminative probes match the gold answers; otherwise, $0$.
    \item \textbf{DCS Score}: $S_{DCS}(G_{j,k}) = 1$ if the VLM judges the image to be consistent with all $n_j$ generated descriptive consistency statements; otherwise, $0$.
    \item \textbf{Joint Score}: $S_{Joint}(G_{j,k}) = S_{VQA}(G_{j,k}) \times S_{DCS}(G_{j,k})$.
\end{itemize}
The performance for prompt $x_j$ is the average Joint Score across its 4 generated images:
$$Score(x_j) = \frac{1}{4} \sum_{k=1}^{4} S_{Joint}(G_{j,k})$$ 
The overall performance on OmniPhys is the mean across all $N=1,551$ prompts:
$$Score_{Final} = \frac{1}{N} \sum_{j=1}^{N} Score(x_j)$$
This hierarchical scoring ensures that the metric reflects the model's consistent ability to manifest correct physics rather than isolated successes.
\begin{figure*}[t]  % [t]=top, [b]=bottom, [h]=here, [!htbp] 更灵活
    \centering
    \includegraphics[width=0.9\linewidth]{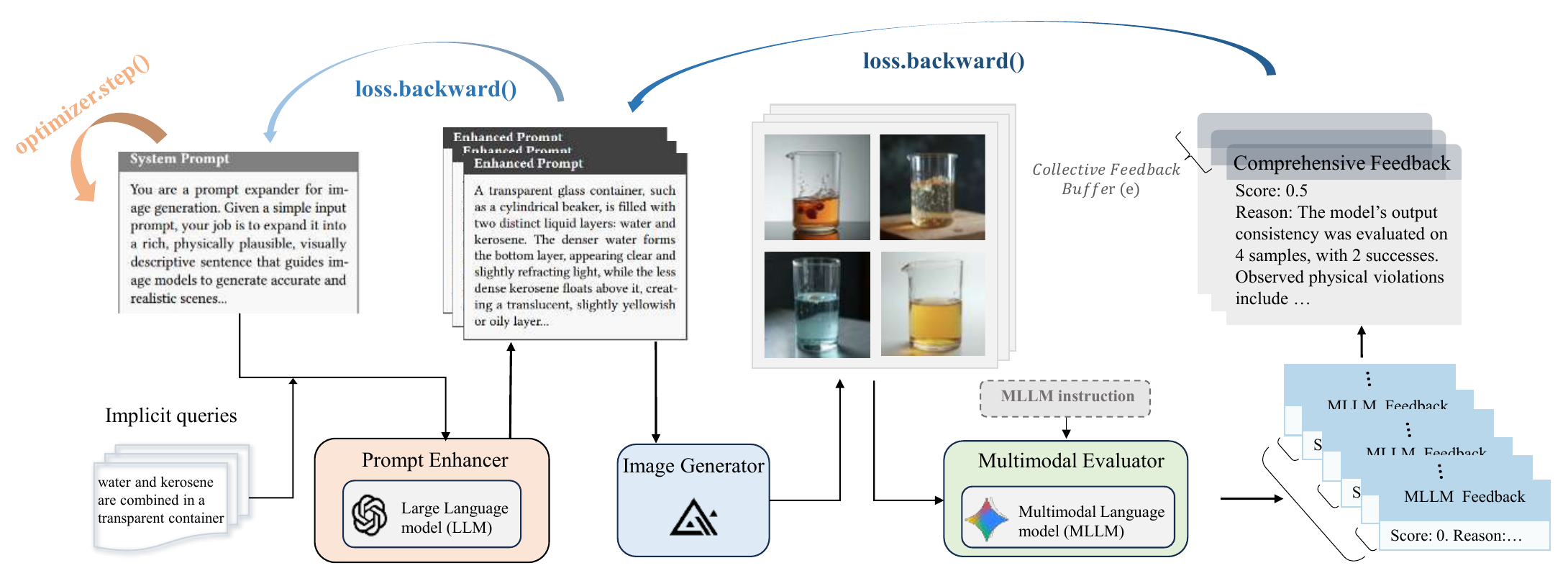}
    \caption{\textbf{The OmniPrompt Framework.} An overview of the iterative alignment loop for physical commonsense in T2I generation. The framework consists of a forward pass for Knowledge-Guided Prompt Enhancement and a backward pass for Meta-Policy Evolution via linguistic gradients.}
    \label{fig:methods}
\end{figure*}

\section{The OmniPrompt Framework}
\label{sec:omniprompt}

In this section, we present \textbf{OmniPrompt}, an iterative framework that treats physical commonsense alignment as a discrete optimization problem over natural-language instructions.
The system establishes a closed-loop pipeline that transforms pixel-level physical violations into structured meta-policy updates through a series of specialized linguistic modules.

\subsection{System Formalization and Optimization Goal}
\label{sec:prelim}

We formalize the alignment process as an optimization of the \textbf{Meta-Policy} $P$, which denotes the high-level system instructions that guide prompt transformation.
The framework consists of five core components:

\begin{enumerate}[leftmargin=*]
    \item \textbf{Prompt Enhancer ($\mathcal{L}$):}
    An LLM-based agent that expands an implicit query $x$ into a physically grounded, visually descriptive prompt $\tilde{x}=\mathcal{L}(x;P)$.
    The meta-policy $P$ defines the enhancement strategy and is the only component updated during training.

    \item \textbf{Stochastic T2I Manifold ($\mathcal{G}$):}
    A generative module that projects the enhanced prompt $\tilde{x}$ into a high-dimensional visual manifold.
    We draw an ensemble of samples from $\mathcal{G}(\tilde{x})$ to capture the stochastic distribution of physical manifestations.

    \item \textbf{Multimodal Evaluator ($\mathcal{E}$):}
    A diagnostic module that verifies each generated image against the query's PKG-grounded DCS probes.
    For every image, $\mathcal{E}$ outputs a binary audit score and a concise rationale, then aggregates them into a collective feedback buffer $e_i$ (Section~\ref{sec:diagnostic-eval}).

    \item \textbf{Feedback Calculator ($\Gamma$):}
    A linguistic derivative operator that propagates batch-level feedback through the TextGrad graph in two stages:
    it first derives a unified prompt-level gradient $g_{\tilde{x}}$ from the concatenated expansions and feedback buffers, then computes sample-wise meta-gradients $\{g_{P}^{(i)}\}_{i=1}^{B}$ by backpropagating through each enhancement branch (Section~\ref{sec:linguistic-gradients}).

    \item \textbf{Meta-Policy Optimizer ($\Phi$):}
    A batch-aware update operator that rewrites $P$ from the aggregated meta-gradients via Textual Gradient Descent (Section~\ref{sec:meta-policy-update}).
\end{enumerate}

\paragraph{Optimization Objective.}
Let $\mathcal{X}$ denote the training prompt pool used for meta-policy optimization.
For each $x_i \in \mathcal{X}$, the enhancer produces $\tilde{x}_i=\mathcal{L}(x_i;P)$, and the evaluator returns a collective feedback buffer $e_i$ with ensemble success rate $\bar{S}_i$ (Section~\ref{sec:diagnostic-eval}).
OmniPrompt seeks the meta-policy
\begin{equation}
    P^* = \arg\max_{P}\; \mathbb{E}_{x_i \sim \mathcal{X}}\big[\bar{S}_i\big].
    \label{eq:opt-goal}
\end{equation}
Only $P$ is updated; the weights of $\mathcal{L}$ and $\mathcal{G}$ remain fixed.

\subsection{Physical Synthesis and Diagnostic Evaluation}
\label{sec:forward-pass}

The forward pass transforms an implicit physical query into a structured diagnostic signal through a sequence of knowledge-driven operations.

\subsubsection{Initialization and Reasoning-before-Synthesis ($\mathcal{L}$)}
\label{sec:reasoning-before-synthesis}

To bridge the gap between abstract physics and visual manifestation, we initialize the meta-policy $P$ with a \textbf{Reasoning-before-Synthesis} paradigm (initial template in Appendix~\ref{app:system_prompt}).
This design ensures that $\mathcal{L}$ infers necessary physical constraints before generating the final description.
Specifically, $\mathcal{L}$ identifies governing laws and predicts object interactions to synthesize a detailed, visually grounded prompt $\tilde{x}$.

\subsubsection{Multi-sample Visual Synthesis ($\mathcal{G}$)}
\label{sec:visual-synthesis}

The generative module $\mathcal{G}$ projects $\tilde{x}$ into the visual domain.
To mitigate stochastic generation noise, we draw an ensemble of $K=4$ independent samples $\mathcal{I}_i=\{I_{i,1},\ldots,I_{i,K}\}$ from $\mathcal{G}(\tilde{x}_i)$ for each enhanced prompt.
This multi-sample design provides a stable empirical basis to distinguish systemic reasoning failures from transient artifacts.

\subsubsection{Physics-aware Multimodal Evaluation ($\mathcal{E}$)}
\label{sec:diagnostic-eval}

The Multimodal Evaluator ($\mathcal{E}$) serves as a diagnostic bridge between visual pixels and linguistic gradients. Expected physical behaviors are specified by the DCS probes associated with each benchmark query, which are generated from the implicit prompt and PKG atomic statements. Unlike methods that rely on a single holistic score~\cite{DBLP:conf/emnlp/PryzantI0L0023}, $\mathcal{E}$ first performs per-image audits and then consolidates them into a collective buffer.

\paragraph{Operational Protocol.} For each image $I_{i,j} \in \mathcal{I}_i$, $\mathcal{E}$ scrutinizes the alignment between visual manifestations and expected physical behaviors, producing a binary score $s_{i,j} \in \{0,1\}$ and a concise rationale $r_{i,j}$:
\begin{equation}
    s_{i,j} =
    \begin{cases}
    1, & \text{if fully consistent with all DCS probes for } x_i, \\
    0, & \text{if any DCS constraint is violated}.
    \end{cases}
    \label{eq:binary-score}
\end{equation}
When $s_{i,j}=0$, the evaluator pinpoints the violation (e.g., ``The object's reflection angle is inconsistent with the light source''), providing the raw diagnostic signal for optimization.

\paragraph{Collective Feedback Construction.}
\emph{Per-query aggregation.} For each implicit query $x_i$, $\mathcal{E}$ consolidates individual audits into a collective feedback buffer
\begin{equation}
    e_i = \langle \bar{S}_i, \mathcal{R}_i \rangle,
    \quad
    \bar{S}_i = \frac{1}{K}\sum_{j=1}^{K} s_{i,j},
    \quad
    \mathcal{R}_i = \bigoplus_{j=1}^{K} r_{i,j}.
    \label{eq:collective-feedback}
\end{equation}
Here, $\bar{S}_i \in [0,1]$ is the ensemble success rate over $K$ images, and $\mathcal{R}_i$ is the consolidated rationale. $\bigoplus$ denotes string concatenation.

\subsection{Meta-Policy Optimization via Linguistic Gradients}
\label{sec:linguistic-gradients}

\emph{Batch-level aggregation.} After processing a batch of $B$ queries through the forward pass, $\Gamma$ and $\Phi$ perform one coupled update on the shared meta-policy $P$. Only the linguistic modules $\mathcal{L}$ and the meta-policy $P$ participate in the TextGrad computation graph.
Image synthesis ($\mathcal{G}$) and multimodal evaluation ($\mathcal{E}$) are external, non-differentiable operators that supply the collective feedback buffer consumed by $\Gamma$.
The backward pass implements a discrete analog of backpropagation over this linguistic graph.

\subsubsection{Feedback Derivation and Joint Differentiation ($\Gamma$)}
\label{sec:gamma}

The Feedback Calculator ($\Gamma$) propagates batch-level feedback through the graph via a two-stage \emph{gradient derivation} procedure.

\paragraph{1. Unified Prompt-level Gradient ($g_{\tilde{x}}$).}
For batch size $B$, $\Gamma$ performs joint batch differentiation over concatenated expansions and feedback buffers:
\begin{equation}
    g_{\tilde{x}} \approx \Gamma\!\left(
        \bigoplus_{i=1}^{B} \tilde{x}_i \;\bigoplus\; \bigoplus_{i=1}^{B} e_i
    \right).
    \label{eq:unified-gradient}
\end{equation}
The resulting unified critique is broadcast to each expansion branch via idempotent concatenation backward in the linguistic graph.
Specifically, $g_{\tilde{x}}$ pinpoints textual flaws that led to the failures reported by $\mathcal{E}$, including ambiguous support descriptions and missing explanations of buoyancy-related mechanisms.

\paragraph{2. Meta-level Gradient Derivation ($\{g_{P}^{(i)}\}$).}
$\Gamma$ subsequently backpropagates $g_{\tilde{x}}$ to the meta-policy $P$.
For each query $i$, it analyzes the causal link between the current instruction $P^{(t)}$ and the corresponding expansion failure:
\begin{equation}
    g_{P}^{(i)} = \nabla_{P} \tilde{x}_i \approx \Gamma\!\left(P^{(t)}, x_i, \tilde{x}_i, g_{\tilde{x}}\right).
    \label{eq:meta-gradient}
\end{equation}
Each meta-gradient diagnoses why $P^{(t)}$ failed to guide the enhancer (e.g., missing explicit requirements for structural equilibrium) and provides strategic suggestions for policy refinement.

\subsubsection{Meta-Policy Evolution and Update ($\Phi$)}
\label{sec:meta-policy-update}

The Meta-Policy Optimizer ($\Phi$) refines $P$ via Textual Gradient Descent through a two-stage \emph{policy update} procedure, using the meta-gradients produced by $\Gamma$.

\paragraph{Gradient Aggregation.}
Each meta-gradient $g_{P}^{(i)}$ is paired with its derivation context and combined into a single textual gradient descent input.
This aggregation consolidates batch-level critiques into one update signal and mitigates conflicting sample-wise edits.

\paragraph{Discrete Policy Update.}
$\Phi$ rewrites $P^{(t)}$ in one step to obtain the refined meta-policy:
\begin{equation}
    P^{(t+1)} = \Phi\!\left(P^{(t)};\, \{g_{P}^{(i)}\}_{i=1}^{B}\right).
    \label{eq:policy-update}
\end{equation}
The update is discrete: $P^{(t+1)}$ is a revised natural-language instruction rather than a continuous parameter change.
Over successive iterations, $P$ gradually incorporates stronger physical constraints, including mechanistic justification and material consistency, which improves alignment across diverse scenarios.

\begin{table}[ht]
\centering
\caption{\textbf{Performance of 12 T2I models on the \texttt{OmniPhys} benchmark} ($N{=}1{,}551$ implicit prompts).
Scores are reported under the dual-path protocol (VQA, DCS, and Joint).
The suffix \textbf{(ep)} denotes the model's native prompt-enhancement configuration (expanded prompt)}
\label{tab:bench_results}
\setlength{\tabcolsep}{4pt} % 减小列间距
\begin{tabular}{@{}lccccc@{}}
\toprule
\textbf{Model} & \textbf{Cat.} & \textbf{Params (B)} & \textbf{VQA} & \textbf{DCS} & \textbf{Joint} \\
\midrule
SD 3.5 Large & Diff & 8 & 0.401 & 0.158 & 0.117 \\
FLUX.1-dev & Diff & 12 & 0.391 & 0.140 & 0.111 \\
HiDream-l1-Full & Diff & 17 & 0.387 & 0.153 & 0.118 \\
Z-Image-Turbo & Diff & 6 & 0.449 & 0.204 & 0.156 \\
Qwen-image & Diff & 27 & 0.377 & 0.167 & 0.131 \\ 
Qwen-image (ep) & Diff & 27 & 0.513 & 0.220 & 0.176 \\
JanusPro-7B & UM & 7 & 0.348 & 0.110 & 0.077 \\
Emu3.5-Image & UM & 34 & 0.469 & 0.216 & 0.161 \\
Lumina-Image 2.0 & AR & 2.6 & 0.371 & 0.131 & 0.102 \\
Infinity & AR & 8 & 0.401 & 0.156 & 0.125 \\
SeedDream-4.0 & Closed & -- & 0.514 & 0.277 & 0.219 \\
Wan2.6-T2I & Closed & -- &  0.491 & 0.266 & 0.212 \\  
Wan2.6-T2I (ep) & Closed & -- & 0.578 & 0.348 & 0.272 \\
Nano Banana Pro & Closed & -- & 0.572 & 0.402 & 0.337 \\
\bottomrule
\end{tabular}
\end{table}

% \begin{table}[ht]
% \centering
% \caption{Performance of Various T2I Models on \texttt{PhysBench}}
% \label{tab:vqa_performance}
% \resizebox{\columnwidth}{!}{% ← 关键：缩放到当前栏宽
% \begin{tabular}{@{}llcccc@{}}
% \toprule
% \textbf{Category} & \textbf{Model} & \textbf{Params (B)} & \textbf{VQA} & \textbf{DCS} & \textbf{Joint} \\
% \midrule
% \multirow{11}{*}{Open-Sourced} 
% & \textit{Diffusion-based} & & & & \\
% & SD 3.5 Large & 8 & 0.401 & 0.158 & 0.117 \\
% & FLUX.1-dev & 12 & 0.391 & 0.140 & 0.111 \\
% & HiDream-l1-Full & 17 & 0.387 & 0.153 & 0.118 \\
% & Z-Image-Turbo & 6 & 0.449 & 0.204 & 0.156 \\
% & Qwen-image (ep) & 27 & 0.513 & 0.220 & 0.176 \\
% \cline{2-6}
% & \textit{Unified Multimodal Model} & & & & \\
% & JanusPro-7B & 7 & 0.352 & 0.323 & 0.113 \\
% & Emu3.5-Image & 34 & 0.469 & 0.216 & 0.161 \\
% \cline{2-6}
% & \textit{Autoregressive-based} & & & & \\
% & Lumina-Image 2.0 & 2.6 & 0.371 & 0.131 & 0.102 \\
% & Infinity & 8 & 0.401 & 0.156 & 0.125 \\
% \midrule
% \multirow{2}{*}{Closed-Sourced} 
% & SeedDream-4.0 & -- & 0.514 & 0.277 & 0.219 \\
% & Nano-Banana-Pro & -- & 0.572 & 0.402 & 0.337 \\
% \bottomrule
% \end{tabular}
% }
% \end{table}
\begin{figure*}[t]  % [t]=top, [b]=bottom, [h]=here, [!htbp] 更灵活
    \centering
    \includegraphics[width=\linewidth, ]{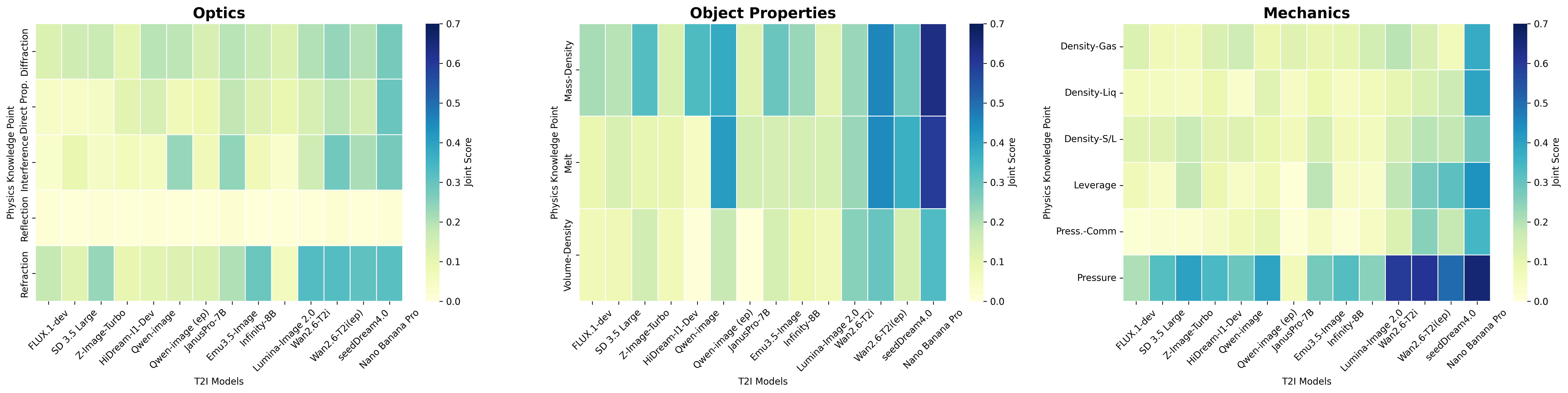}
    \caption{Granular performance across 14 PKPs. Heatmaps show Joint Scores across Optics (left), Object Properties (center), and Mechanics (right). Darker shades represent higher consistency, highlighting systemic bottlenecks like \textit{Reflection} and \textit{Leverage Principle} across all architectures.}
    \label{fig:benchmark_result_analysis}
\end{figure*}

\begin{figure}[t]  % [t]=top, [b]=bottom, [h]=here, [!htbp] 更灵活
    \centering
    \includegraphics[width=\linewidth]{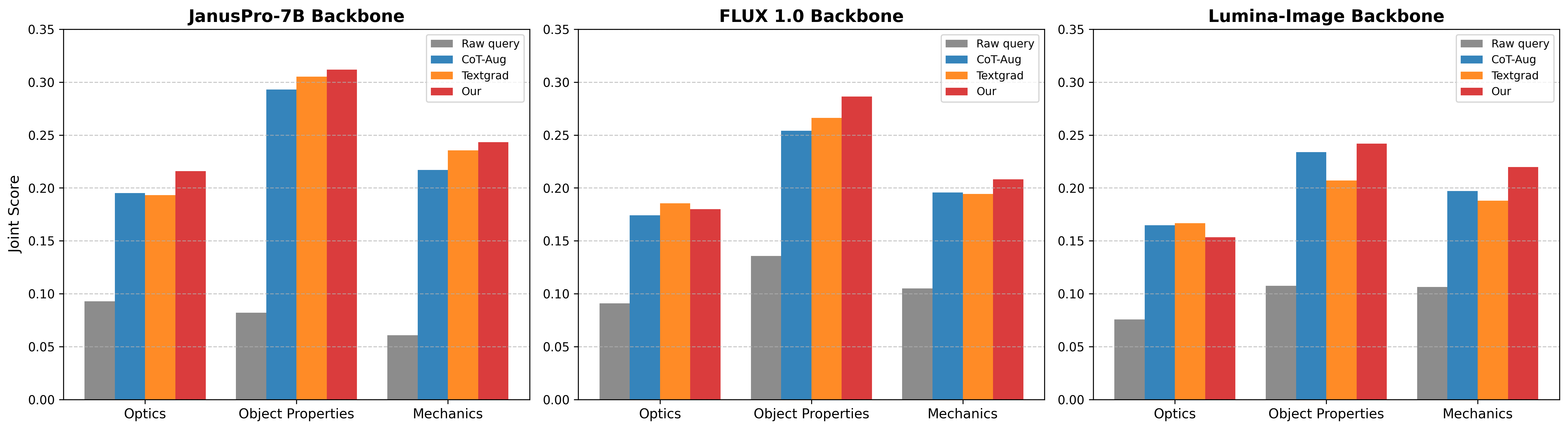}
    \caption{Performance gains across physical domains. Grouped bar charts show Joint Score improvements of \texttt{OmniPrompt} over baselines in Optics, Object Properties, and Mechanics.}
    \label{fig:domain_Radar_Chart}
\end{figure}

\begin{figure}[t]  % [t]=top, [b]=bottom, [h]=here, [!htbp] 更灵活
    \centering
    \includegraphics[width=\linewidth]{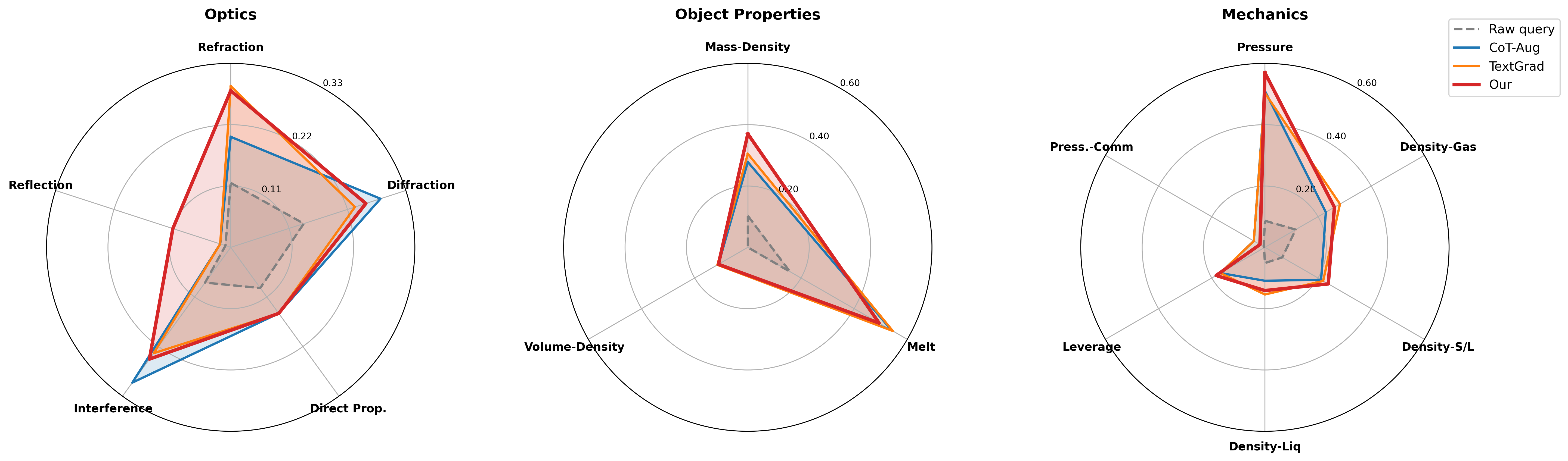}
    \caption{Granular knowledge point recovery on the JanusPro-7B backbone. This radar chart compares the alignment of \texttt{OmniPrompt}, TextGrad, and CoT-Aug across 14 PKPs.}
    \label{fig:Janus_Bar_Chart}
\end{figure}

\section{Experiments}
\label{sec:experiments}
\subsection{Experimental Setup}
\label{sec:experimental_setup}
\subsubsection{OmniPhys Taxonomy}
\label{sec:omniphys-taxonomy}

Our benchmark, \texttt{OmniPhys}, comprises $N=1{,}551$ implicit prompts anchored to 14 PKPs from our PKG.
As illustrated in Figure~\ref{fig:physbench_stats}, samples are organized under three domains:
\begin{itemize}[leftmargin=*]
    \item \textbf{Mechanics (55.8\%):}
    Covers fluid--structure interactions and force equilibrium, including density-driven buoyancy (solid--fluid, gas, and immiscible liquids), leverage, and pressure (including force-induced deformation and communicating vessels).

    \item \textbf{Optics (24.5\%):}
    Evaluates ray and wave optics, including linear propagation, reflection, refraction, diffraction, and interference.

    \item \textbf{Object Properties (19.7\%):}
    Tests intrinsic material relations, including mass--volume--density consistency and phase change.
\end{itemize}

\subsubsection{Evaluation Landscape: T2I Models}
We benchmark our framework against 12 representative T2I models to ensure a robust cross-architecture assessment:
\begin{itemize}
    \item \textbf{Open Source}: We select models representing three synthesis paradigms: \textbf{Diffusion-based} (FLUX.1-dev \cite{DBLP:journals/corr/abs-2506-15742}, SD-3.5-Large \cite{DBLP:conf/icml/EsserKBEMSLLSBP24}, Z-Image-Turbo \cite{DBLP:journals/corr/abs-2511-22699}, HiDream-I1-Full \cite{DBLP:journals/corr/abs-2505-22705}, and Qwen-image \cite{DBLP:journals/corr/abs-2508-02324}), \textbf{Unified Multimodal} (JanusPro-7B \cite{DBLP:journals/corr/abs-2501-17811}, Emu3.5-Image \cite{DBLP:journals/corr/abs-2510-26583}), and \textbf{Autoregressive} (Infinity-8B \cite{DBLP:conf/cvpr/HanL0YZYPL25}, Lumina-Image-2.0 \cite{DBLP:journals/corr/abs-2503-21758}).
    
    \item \textbf{Closed source}: We include three leading commercial engines, Wan2.6-T2I \cite{DBLP:journals/corr/abs-2503-20314}, SeedDream-4.0 \cite{DBLP:journals/corr/abs-2509-20427}, and Gemini 3 Pro Image \cite{google2025gemini3proimage} (``Nano Banana Pro'') to represent the industrial upper bound of generative performance.
\end{itemize}

\subsubsection{Baseline Configurations}
We establish three baselines to isolate the performance gains from our collective optimization strategy:
\begin{itemize}
    \item \textbf{Raw Query}: Direct generation using original implicit prompts $x$ from OmniPhys.
    
    \item \textbf{Zero-shot CoT-Aug (CoT-Aug)}: Enhancing prompts $x$ using the initial meta-policy $P^{(0)}$ without any iterative updates.
    
    \item \textbf{Instance-level Optimization (TextGrad)}: An iterative baseline where the meta-policy is updated using independent, sample-wise gradients.
\end{itemize}

\subsubsection{Implementation Details}

\paragraph{Models and optimization targets.}
We use GPT-4o (2024-11-20)~\cite{openai2024gpt4oapi} as the prompt enhancer, TextGrad backward engine, and meta-policy optimizer; Gemini-2.5-Pro~\cite{DBLP:journals/corr/abs-2507-06261} as the physics-aware multimodal evaluator. To assess cross-model transfer, evolved policies are evaluated on FLUX.1-dev~\cite{DBLP:journals/corr/abs-2506-15742}, JanusPro-7B, and Lumina-Image-2.0~\cite{DBLP:journals/corr/abs-2503-21758}.

\paragraph{Data splits and training pool.}
OmniPhys contains 1,551 prompts in total.
For experiments, we use a PKP-balanced subset of 1,387 prompts (train/val/test = 274/274/839; ratio 1:1:3), obtained by downsampling \textit{diffraction of light} from 224 to 60.
Meta-policy optimization runs on a fixed pool of 37 training prompts ($\sim$2--3 per PKP); validation and test are reserved for rollback and final reporting.

\paragraph{Evaluation and optimization protocol.}
During training, $\mathcal{E}$ provides DCS-only feedback on the 37 prompt pool; validation and test use the full OmniPhys Dual-Path protocol.
Each query uses $K=4$ generated images.
TextGrad updates are triggered every $B$ accumulated queries; we study $B \in \{2,4,6,8,10,12\}$.
After each update, the candidate policy is validated on the full validation split; we revert to the previous best policy if the Joint Score does not improve, and stop after 5 consecutive failures (patience = 5, up to 5 epochs). An example of End-to-end illustration ($B=2$, $K=4$) is found in the Appendix~\ref{app:running-example}.
\subsection{Benchmarking Results on \texttt{OmniPhys}}
\label{subsec:benchmarking_results}
Table~\ref{tab:bench_results} evaluates 12 T2I models on the full \texttt{OmniPhys} ($N{=}1{,}551$) using VQA, DCS, and Joint scores. We observe a persistent gap where \textbf{VQA scores significantly exceed DCS scores}, suggesting that models can pass binary visual probes via simple heuristics but struggle to satisfy the finer descriptive consistency constraints required by DCS.
Consequently, we adopt the \textbf{Joint Score} as the primary metric to mitigate this evaluation bias.
Overall scores remain relatively low even for strong systems, which is consistent with the strict all-or-nothing Joint criterion on implicit physical queries. By synthesizing the macro trends from Table~\ref{tab:bench_results} with the PKP-level heatmaps in Figure~\ref{fig:benchmark_result_analysis}, we derive three key observations regarding the current T2I landscape:

\paragraph{Frontier Models and Domain-Specific Limits.}
Closed-source models (e.g., \textbf{Nano Banana Pro}, \textbf{Wan2.6-T2I (ep)}) achieve the strongest overall performance in physical alignment.
However, this superiority is highly domain-specific.
As shown in Figure~\ref{fig:benchmark_result_analysis}, while commercial models excel in \textit{Pressure} ($>0.60$), they encounter a persistent bottleneck in \textit{Optics}, where even top-tier industrial engines achieve near-zero scores on PKPs such as \textit{Reflection of light}.

\paragraph{Limited Gains from Model Scale.}
Increased parameter size does not guarantee a breakthrough in physical reasoning.
Large-scale models like \textbf{Emu3.5 (34B)} show no transformative advantage over mid-sized counterparts.
Scaling primarily improves ``texture-based'' physics (e.g., material surfaces) but fails on PKPs requiring global mechanical constraints, such as \textit{Leverage Principle}, which exhibit consistently low scores across models.
This confirms that raw capacity cannot substitute for explicit structural priors in prompt formulation.

\paragraph{Prompt Expansion as a Material Catalyst.}
Prompt expansion effectively unlocks latent potential by providing explicit physical context.
Significant gains are observed in \textbf{Object Properties} PKPs (e.g., \textit{Melt}, \textit{Object properties--mass and density}), as evidenced by the performance leap from \textbf{Wan2.6-T2I} to its enhanced version \textbf{(ep)}.
These detailed descriptions of volume and state changes help bridge the gap between abstract semantic intent and physical visual synthesis.
However, expansion offers minimal improvement on logic-intensive \textit{Optics} PKPs, supporting our motivation to optimize meta-policies for reasoning-oriented enhancement.

\begin{table}[t]
\centering
\caption{\textbf{Main results of \texttt{OmniPrompt} across T2I backbones.}
Evaluated on the PKP-balanced split.}
\label{tab:omniprompt_results}
\begin{adjustbox}{max width=\linewidth}
\begin{tabular}{llccc}
\toprule
\textbf{Backbone Model} & \textbf{Strategy} & \textbf{VQA} & \textbf{DCS} & \textbf{Joint} \\
\midrule
\multirow{4}{*}{FLUX.1-dev} & Raw Query & 0.389 & 0.132 & 0.110 \\
 & CoT-Aug & 0.478 & 0.252 & 0.205 \\
 & TextGrad & 0.465 & 0.260 & 0.209 \\
 & \textbf{OmniPrompt (Ours)} & \textbf{0.478} & \textbf{0.281} & \textbf{0.221} \\
\midrule
\multirow{4}{*}{JanusPro-7B} & Raw Query & 0.333 & 0.100 & 0.071 \\
 & CoT-Aug & 0.499 & 0.305 & 0.230 \\
 & TextGrad & 0.522 & 0.321 & 0.244 \\
 & \textbf{OmniPrompt (Ours)} & \textbf{0.523} & \textbf{0.322} & \textbf{0.254} \\
\midrule
\multirow{4}{*}{Lumina-Image 2.0} & Raw Query & 0.365 & 0.126 & 0.102 \\
 & CoT-Aug & 0.465 & 0.243 & 0.200 \\
 & TextGrad & 0.469 & 0.234 & 0.189 \\
 & \textbf{OmniPrompt (Ours)} & \textbf{0.472} & \textbf{0.266} & \textbf{0.214} \\
\bottomrule
\end{tabular}
\end{adjustbox}
\end{table}

\subsection{Effectiveness of \texttt{OmniPrompt}}
\label{subsec:omniprompt_effectiveness}

Table~\ref{tab:omniprompt_results} and Figure~\ref{fig:domain_Radar_Chart} summarize the performance of \texttt{OmniPrompt} across three representative T2I architectures.
By benchmarking against \textbf{Raw Query}, \textbf{CoT-Aug}, and \textbf{TextGrad}, we evaluate the meta-policy's efficacy in addressing complex physical constraints.

\paragraph{Optimization Dynamics and Stability.}
\texttt{OmniPrompt} consistently establishes the highest performance ceiling among compared methods, achieving Joint Scores of 0.221, 0.254, and 0.214 for FLUX.1-dev, JanusPro-7B, and Lumina-Image-2.0, respectively.
Comparison with TextGrad highlights the benefit of batch-merged optimization. While single-query, instance-level backward passes are prone to such gradient hallucinations, OmniPrompt aggregates per-query feedback over $B$ queries before updating $P$, yielding batch-level diagnostics that are less sensitive to seed- and query-local noise (Appendix~\ref{app:raw_trace}). This robustness is particularly evident on Lumina-Image-2.0, where TextGrad's sample-wise sensitivity causes it to underperform even the simpler CoT-Aug baseline.

\paragraph{Physical State Manifestation.}
Structured physical descriptions (CoT-Aug) significantly improve DCS scores compared to raw queries, notably surging from 0.100 to 0.305 on JanusPro-7B.
This suggests that explicit descriptive consistency constraints provide a necessary foundation for models to move from simple object recognition toward physically checkable execution.
\texttt{OmniPrompt} further refines these descriptions through meta-policy evolution, yielding more consistent manifestations of force and material interactions.

\paragraph{Domain-Specific Knowledge Recovery.}
As shown in Figures~\ref{fig:domain_Radar_Chart} and~\ref{fig:Janus_Bar_Chart}, the evolved meta-policy is particularly effective on \textbf{Mechanics} and \textbf{Object Properties} PKPs.
\texttt{OmniPrompt} successfully improves challenging cases such as \textit{Leverage Principle} and \textit{Pressure-communicator}, which simple text expansion fails to resolve.
For material-centric PKPs, it guides the model toward more verifiable visual evidence, such as puddle formation for \textit{Melt} or clearer volume--density relations for mass--density PKPs.
While \textit{Reflection of light} remains difficult, we observe meaningful progress on other \textit{Optics} PKPs, including \textit{Refraction of light} and \textit{Diffraction of light}, after instruction refinement for visual displacement and fringe patterns.

\begin{figure}[t]
    \centering
    \includegraphics[width=0.85\linewidth]{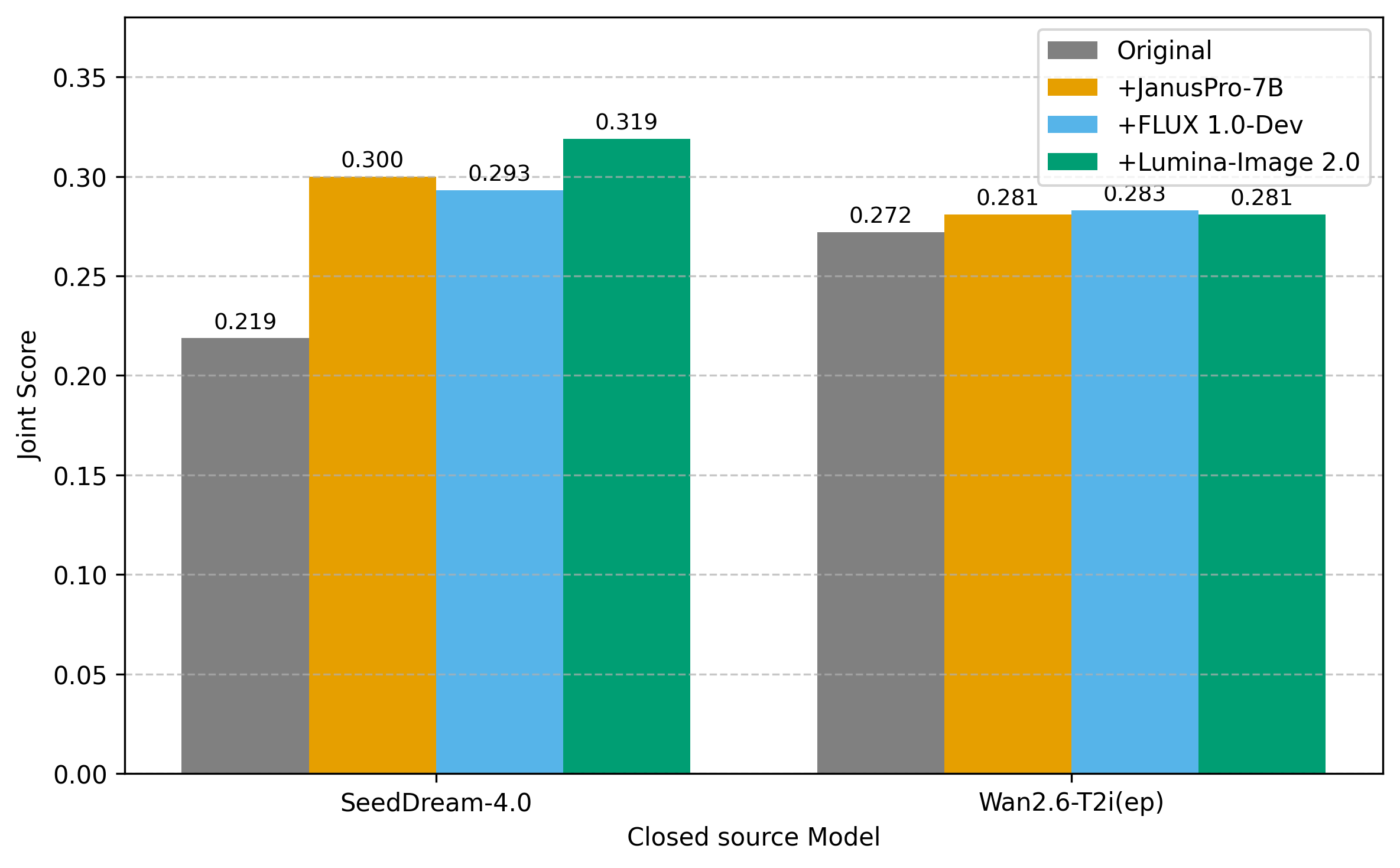}
    \caption{Cross-Model Policy Generalization. Results showing the transfer of meta-policies evolved on open-source backbones to closed-source engines.}
    \label{fig:transformer_prompt}
\end{figure}

\begin{table}[t]
\centering
\caption{Human evaluation of VQA-DCS probe quality and VLM scoring alignment. For 50 randomly sampled VQA-DCS probes, three physics-trained annotators assessed four criteria (validity rates and Gwet's AC1 coefficients are shown). For 200 generated images, we report the human-VLM agreement rate and inter-annotator reliability (Gwet's AC1).}
\label{tab:human_eval}
\begin{tabular}{lcc}
\toprule
\textbf{Evaluation aspect} & \textbf{Validity / Agree.} & \textbf{Gwet's AC1} \\
\midrule
\multicolumn{3}{c}{\textbf{VQA-DCS probes (N=50)}} \\
\quad VQA focus \& corr. & 98.0\% & 0.896 \\
\quad DCS physical corr. & 96.0\% & 0.892 \\
\quad DCS visual feas. & 96.0\% & 0.892 \\
\quad VQA-DCS sem. cons. & 100.0\% & 0.986 \\
\midrule
\multicolumn{3}{c}{\textbf{Image labeling (N=200)}} \\
\quad Human-VLM agreement & 80.0\% & --- \\
\quad Inter-annotator agree. & 65.8\% (3-way) & 0.637 \\
\bottomrule
\end{tabular}
\end{table}

\subsection{Human Evaluation}

To validate the reliability of our automated evaluation pipeline and address concerns about evaluator bias, we conducted a human study with three physics-trained annotators.
First, we randomly selected 50 LLM-generated VQA--DCS probe pairs anchored to PKG PKPs and evaluated each on four binary criteria: (i)~\textbf{VQA focus \& correctness}: whether the VQA question targets the core physical phenomenon and its answer is physically correct; (ii)~\textbf{DCS physical correctness}: whether the descriptive consistency statement conforms to established physical laws; (iii)~\textbf{DCS visual feasibility}: whether the described state can be unambiguously observed in a static 2D image; and (iv)~\textbf{VQA--DCS consistency}: whether the VQA question and the DCS description refer to the same underlying physical state.
Second, we assessed the alignment between VLM scoring and human judgment on 200 generated images (50 prompts $\times$ 4 images from FLUX.1-dev).

As shown in Table~\ref{tab:human_eval}, all four probe criteria achieved high validity rates (96\%--100\%) with substantial to almost perfect inter-annotator agreement (Gwet's AC1: 0.892--0.986).
On the 200 generated images, the VLM agreed with human majority at 80.0\%, and overall annotator consensus was moderate-to-substantial (AC1=0.637).
These results confirm the physical soundness of our benchmark probes and the reliability of VLM-based scoring.

\subsection{Robustness and Generalization}
\label{subsec:robustness}

We evaluate the stability and generalizability of our framework by analyzing optimization trajectories and cross-model transferability.

\paragraph{Collective Feedback and Stability.}
As evidenced in our sensitivity study over batch size $B \in \{2,4,6,8,10,12\}$, OmniPrompt maintains a robust performance plateau for JanusPro-7B (0.239--0.254), whereas instance-level TextGrad optimization exhibits volatile fluctuations. This stability is primarily attributed to \emph{batch-level} collective feedback: linguistic gradients are derived only after merging $B$ per-query buffers, which aggregates batch-level linguistic gradients to filter stochastic image artifacts and surface shared physical misconceptions. A comparative diagnostic trace of TextGrad versus OmniPrompt gradients is provided in Appendix~\ref{app:raw_trace}.

\paragraph{Cross-Model Policy Generalization.}
We assess the transferability of evolved policies to unseen closed-source engines.
As illustrated in Figure~\ref{fig:transformer_prompt}, meta-policies developed on open-source backbones yield consistent performance gains across diverse targets.
Notably, SeedDream-4.0 achieves a 45.7\% relative gain in Joint Score when guided by the Lumina-evolved policy.
The fact that policies from various backbones yield uniformly high gains suggests that our optimization captures transferable physical reasoning patterns rather than backbone-specific wording alone.
These results support the use of the evolved meta-policy as a physics-aware prompt enhancement across diverse generative paradigms.
\section{Conclusion}
This paper introduces \textbf{OmniPhys}, a benchmark grounded in a PKG to diagnose reasoning deficits in T2I models. Our evaluation reveals systemic bottlenecks in structural and optical consistency, where models prioritize statistical pixel correlations over physical laws. To address these failures, we propose \textbf{OmniPrompt}, an iterative framework that distills collective feedback to evolve robust and "physics-aware" meta-policies. Experimental results demonstrate that our approach significantly enhances physical alignment and generalizes across diverse generative backbones. 
%% The next two lines define the bibliography style to be used, and
%% the bibliography file.
\begin{acks}
This work is funded by National Natural Science Foundation of China (NSFCU23B2055/NSFC62306276), New Generation Artificial Intelligence-National Science and Technology Major Project 2030 (2025ZD0122800), Yongjiang Talent Introduction Programme (2022A-238-G), and Fundamental Research Funds for the Central Universities (226-2023-00138). This work was supported by Ant Group.
\end{acks}
\bibliographystyle{ACM-Reference-Format}
\bibliography{main}

\appendix
\section{Implementation of System Prompt Initialization}
\label{app:system_prompt}

\begin{tcolorbox}[
    title={System Prompt Specification for $\mathcal{L}$},
    colback=gray!5,
    colframe=gray!50!black,
    fonttitle=\bfseries,
    sharp corners,
    boxrule=1pt,
    left=5pt,
    right=5pt,
    top=5pt,
    bottom=5pt,
    arc=3pt,
    boxsep=2pt,
    before skip=10pt,
    after skip=10pt
]

\noindent\textbf{Role:} You are a prompt expander for image generation. Given a simple input prompt, your job is to expand it into a rich, physically plausible, visually descriptive sentence that guides image models to generate accurate and realistic scenes.

\vspace{0.5em}

\noindent\textbf{Step 1: Physical Reasoning} \\
Reason about the implicit physical states of objects in the prompt. Think through what physical principles are involved and how they influence the object’s behavior.

\vspace{0.5em}

\noindent\textbf{Step 2: Generate the expanded descriptive text.} \\
Combine the original prompt with the physical insights from Step 1 to produce a rich, coherent, and visually specific description—optimized for image generation.

\vspace{0.5em}

Finally, output your expanded text in JSON format:
\{"expanded\_text": "The final expanded text for image generation — descriptive, concrete, and physics-aware."\}
\end{tcolorbox}

\section{Stability Analysis of Collective Linguistic Gradients}
\label{app:gradient_stability}
\begin{figure}[htbp]
    \centering
    \includegraphics[width=0.95\linewidth]{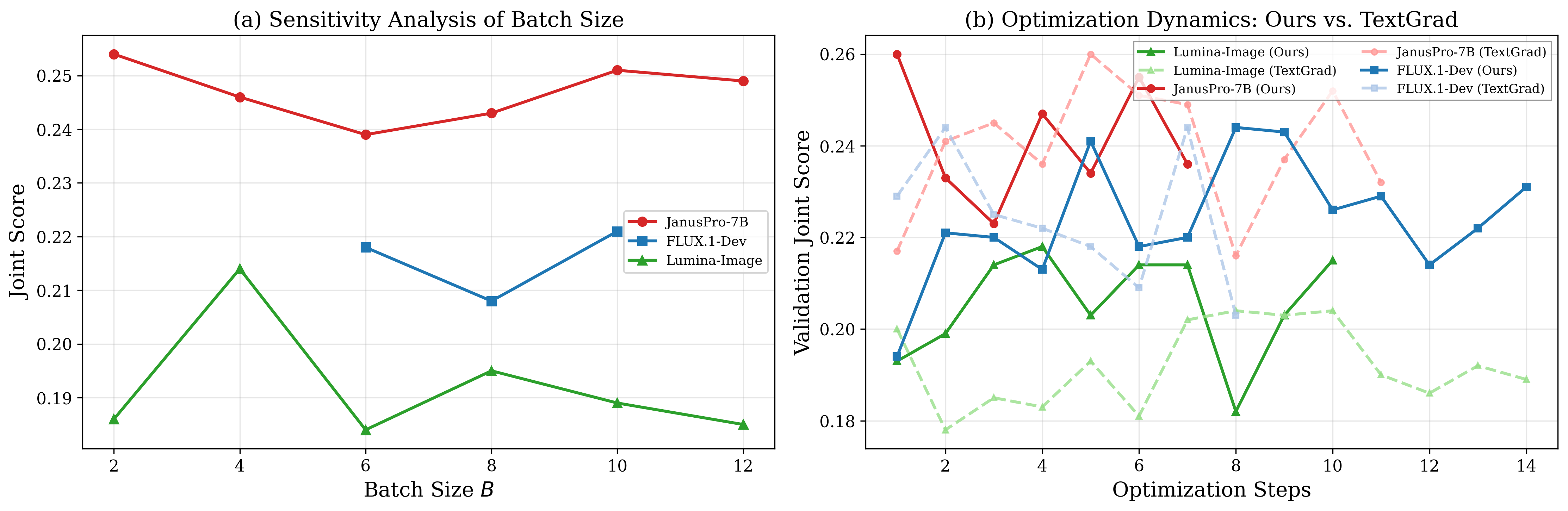}
    \caption{Sensitivity analysis and optimization dynamics. (a) Joint Scores across different batch sizes $B$. (b) Step-wise validation trajectories comparing our collective optimization against TextGrad.}
    \label{fig:batch_size}
\end{figure}

As illustrated in Figure~\ref{fig:batch_size}, the scale of collective feedback $B$ plays a critical role in balancing the precision of linguistic gradients with computational efficiency. In Figure~\ref{fig:batch_size} (a), we observe that JanusPro-7B is highly robust to variations in collective scale, suggesting a relatively smooth optimization landscape. In contrast, FLUX.1-dev exhibits a sharp performance drop outside the $B \in [6, 10]$ range, indicating that large-scale Diffusion Transformers (DiTs) are more susceptible to the density of meta-level linguistic gradients. Furthermore, Figure~\ref{fig:batch_size} (b) highlights the anti-noise capability inherent in our collective feedback mechanism; while instance-level methods (e.g., TextGrad) suffer from stochastic oscillations triggered by outlier images in the T2I manifold, \texttt{OmniPrompt} leverages cross-sample consensus to maintain a steady upward trajectory toward physical alignment.

\section{Illustration of PKG-Guided Coupled Probe Generation}
\label{app:probe-generation}

For each implicit query $x$, OmniPhys retrieves the anchor PKP's atomic physical statements and prompts an LLM with a visual-physical template, and PKP-aligned few-shot demonstrations.
The LLM selects $n$ scene-specific physical state points and co-generates coupled units $u_i=(\text{VQA}_i, \text{DCS}_i)_{i=1}^{n}$, where DCS denotes a Descriptive Consistency Statement (Section~\ref{sec:dual_path_protocol}).

\begin{tcolorbox}[
    colback=white, 
    colframe=black!70, 
    arc=2mm, 
    boxsep=4pt,
    left=5pt, 
    right=5pt, 
    top=3pt, 
    bottom=3pt,
    % 关键修改：用 {} 包裹整个 title 内容，保护内部的逗号
    title={\textbf{Case: Reflection in a Flat Mirror} \quad (\textit{Reflection of light}, $n=2$)}
]

\small
\textbf{Input.}\;
$x$:\ \texttt{``A cylindrical candle sits on a table facing a wall-mounted mirror.''}\;
Atomic statements: equal incidence/reflection angles; object--image equidistance to the mirror.

\vspace{0.4em}
\textbf{Generated units.}
\begin{itemize}[leftmargin=1.2em, nosep, itemsep=3pt]
    \item \textbf{Unit 1 (size):}\;
    VQA: ``Is the reflected candle smaller than the actual candle?''\ \textbf{Ans:} no;\;
    DCS: same height and width in object and reflection.
    \item \textbf{Unit 2 (distance):}\;
    VQA: ``Is the reflection closer to the mirror than the candle?''\ \textbf{Ans:} no;\;
    DCS: reflection is equidistant behind the mirror; candle-to-reflection distance is twice the object-to-mirror distance.
\end{itemize}
\end{tcolorbox}

\section{Diagnostic Trace of Evolved Linguistic Gradients}
\label{app:raw_trace}

Table~\ref{tab:raw_trace_comparison} contrasts the gradient-update behavior of \textbf{TextGrad} versus \textbf{OmniPrompt}.
Both runs share the same initial meta-policy $P^{(0)}$, evaluator $\mathcal{E}$, and ensemble size $K=4$, and include the same gas-density queries (e.g., a helium-filled inflatable toy in CO$_2$ and an SF$_6$-filled plastic bag in H$_2$).

\textbf{TextGrad} performs \emph{single-query} backward: each query is evaluated and backpropagated separately, producing query-local $g_{\tilde{x}}$ and $g_P$ before $\Phi$ rewrites $P$.
\textbf{OmniPrompt} performs \emph{batch-merged} backward with $B=2$: forward passes accumulate two queries, merged $(\tilde{x}_i,e_i)$ yield one unified $g_{\tilde{x}}$, two branch-specific meta-gradients $\{g_{P}^{(i)}\}_{i=1}^{2}$, and a single $\Phi$ update.
Stages~1--3 below denote $g_{\tilde{x}}$, meta-gradients, and $P^{(t+1)}$ (Eqs.~\ref{eq:unified-gradient}--\ref{eq:policy-update}).

\begin{table*}[tbp]
\centering
\caption{TextGrad vs.\ OmniPrompt: representative linguistic-gradient traces (excerpts abbreviated from our logs).}
\label{tab:raw_trace_comparison}
\small
\renewcommand{\arraystretch}{1.22}
\begin{tabularx}{\textwidth}{>{\raggedright\arraybackslash}X | >{\raggedright\arraybackslash}X}
\toprule
\textbf{TextGrad (single-query backward)} & \textbf{OmniPrompt ($B=2$ batch merge)} \\
\midrule

\textbf{Stage 1: $g_{\tilde{x}}$} &
\textbf{Stage 1: unified $g_{\tilde{x}}$} \\
\textit{Excerpt:} ``The primary issue is a dense plastic bag floating in mid-air, violating gravity and buoyancy\ldots\ suggest tethering, external support, or a lighter-than-air fill.'' &
\textit{Excerpt:} One merged critique spans the batch: (i)~``inflatable toy floating in mid-air without visible support or buoyancy mechanism''; (ii)~vague melt/heat wording that should specify concrete transformation (e.g., softening, bubbling). \\
\textit{Analysis:} Feedback is tied to \textbf{one query's} dominant failure at a time. &
\textit{Analysis:} Batch merge produces \textbf{one} $g_{\tilde{x}}$ covering multiple failure modes across the two queries. \\

\midrule
\textbf{Stage 2: $g_P$} &
\textbf{Stage 2: $\{g_{P}^{(i)}\}_{i=1}^{2}$} \\
\textit{Excerpt:} ``Add explicit emphasis on material and physical properties\ldots\ describe material state (gas/liquid/solid) and container interaction.'' &
\textit{Excerpt:} ``Strengthen Reason-before-Synthesis\ldots\ ensure physical laws hold and \textbf{state the mechanism} (e.g., buoyancy, suspension).'' (two branch critiques aggregated by $\Phi$.) \\
\textit{Analysis:} Meta-gradient mainly adds \textbf{content-level} directives. &
\textit{Analysis:} Meta-gradient targets the \textbf{reasoning procedure} in $P$. \\

\midrule
\textbf{Stage 3: $P^{(t+1)}$} &
\textbf{Stage 3: $P^{(t+1)}$} \\
\textit{Excerpt:} 5 steps: analyze states $\rightarrow$ environment $\rightarrow$ generate $\rightarrow$ verify $\rightarrow$ refine iteratively. &
\textit{Excerpt:} 8 steps: physical principles $\rightarrow$ material $\rightarrow$ environment $\rightarrow$ cause--effect $\rightarrow$ visual cues $\rightarrow$ edge cases $\rightarrow$ review $\rightarrow$ JSON. \\
\textit{Analysis:} Policy grows via appended rules after query-local failures. &
\textit{Analysis:} Policy becomes longer but more \textbf{explicitly staged} (principles, materials, validation). \\
\bottomrule
\end{tabularx}
\end{table*}

\section{End-to-End Illustration of OmniPrompt}
\label{app:running-example}

This appendix walks through \textbf{once OmniPrompt update} with $B=2$ and $K=4$, complementing Section~\ref{sec:omniprompt}.
External modules: $\mathcal{G}$ (JanusPro-7B), $\mathcal{E}$ (Gemini-2.5-Pro); graph participants: $\mathcal{L}$, $P$, $\Gamma$, $\Phi$ (GPT-4o).
Full instruction templates and the complete training log are in our code release.

\paragraph{Fixed templates (abbreviated).}
\textbf{Meta-policy $P^{(0)}$.}
Reasoning-before-Synthesis: infer physical states, then emit JSON \texttt{\{"expanded\_text": ...\}}.
\textbf{Evaluator $\mathcal{E}$ (training).}
VLM compares each image against the query's \emph{pre-generated DCS probes}; output format:
\texttt{Score: [0|1]. Reason: [one sentence]}.
Per-image scores are aggregated into $e_i=\langle \bar{S}_i, \mathcal{R}_i \rangle$ (Eq.~\ref{eq:collective-feedback}).
\textbf{$\Gamma$ and $\Phi$.}
Standard TextGrad string-function backward and Textual Gradient Descent; batch loss binds
$\bigoplus_i \tilde{x}_i$ to merged feedback, yielding one $g_{\tilde{x}}$ and $\{g_{P}^{(i)}\}_{i=1}^{B}$, then one rewrite of $P$.
\subsection{Running Example ($B=2$, $K=4$)}
\label{app:stages}

\begin{center}
\small
\begin{tabular}{clp{0.50\linewidth}}
\toprule
\textbf{Stage} & \textbf{Module} & \textbf{Output} \\
\midrule
I   & $\mathcal{L}$ &
\textbf{$i=1$}: $x_1$ (helium toy in sealed CO$_2$ room) $\rightarrow$ $\tilde{x}_1$. \\
    &                 &
\textbf{$i=2$}: $x_2$ (SF$_6$ bag in sealed H$_2$ room) $\rightarrow$ $\tilde{x}_2$. \\
\midrule
II  & $\mathcal{G},\mathcal{E}$ &
For each $i\in\{1,2\}$: sample $K=4$ images and form
$e_i=\langle \bar{S}_i,\mathcal{R}_i\rangle$ (e.g., $\bar{S}_1=0.50$ with unsupported-hovering violations). \\
\midrule
III & loss node &
Batch merge: $\bigoplus_{i=1}^{2}\tilde{x}_i$ and $\bigoplus_{i=1}^{2} e_i$ $\rightarrow$ TextGrad root. \\
\midrule
IV  & $\Gamma$ &
One unified $g_{\tilde{x}}$; two meta-gradients $g_P^{(1)}, g_P^{(2)}$ (one per query branch). \\
\midrule
V   & $\Phi$ &
Aggregate $\{g_P^{(1)}, g_P^{(2)}\}$ $\rightarrow$ single update $P^{(t+1)}$. \\
\bottomrule
\end{tabular}
\end{center}

\end{document}